%% file: main.tex
\documentclass[10pt,twocolumn,letterpaper]{article}

\usepackage[pagenumbers]{iccv} 
\usepackage{fancyvrb}
\usepackage{listings}
\usepackage{xcolor}
\usepackage{fancyvrb}
\usepackage[pagebackref,breaklinks,colorlinks,allcolors=iccvblue]{hyperref}

\definecolor{iccvblue}{rgb}{0.21,0.49,0.74}

\usepackage{xcolor}
\usepackage{fancyvrb}

\usepackage[pagebackref,breaklinks,colorlinks,allcolors=iccvblue]{hyperref}

\def\paperID{11441} 
\def\confName{ICCV}
\def\confYear{2025}

\title{What Changed and What Could Have Changed? 
State-Change Counterfactuals for Procedure-Aware Video Representation Learning}

\author{
Chi-Hsi Kung$^*$ \\
Indiana University \\
{\tt\small kung@iu.edu}
\and
Frangil Ramirez$^*$ \\
Indiana University \\
{\tt\small fraramir@iu.edu}
\and
Juhyung Ha \\
Indiana University \\
{\tt\small yuhha@iu.edu}
\and
Yi-Ting Chen$^\dagger$ \\
National Yang-Ming Chiao-Tung University \\
{\tt\small ychen@cs.nycu.edu.tw}
\and
David J. Crandall$^\dagger$ \\
Indiana University \\
{\tt\small djcran@iu.edu}
\and
Yi-Hsuan Tsai$^\dagger$ \\
Atmanity Inc. \\
{\tt\small wasidennis@gmail.com}
}
\begin{document}

\twocolumn[{%
    \renewcommand\twocolumn[1][]{#1}%
    \maketitle
    \begin{center}
        \centering
        \captionsetup{type=figure}
        \includegraphics[width=1.0\textwidth]{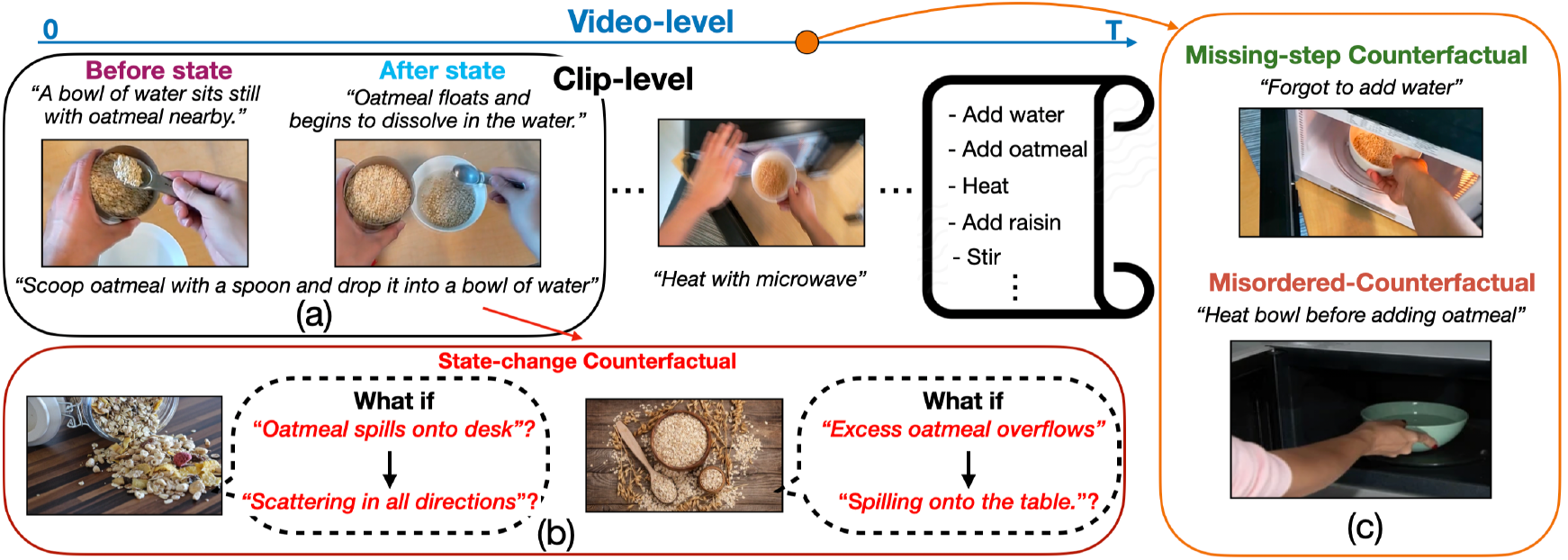} 
        \captionof{figure}{
            Illustration of ``what changed'' (actual action-induced transformations) and ``what could have changed'' (hypothetical deviations) for both clip- and video-level procedures. (a) The clip narration \emph{``Scoop oatmeal with a spoon and drop it into a bowl of water''} changes the scene's \textcolor{RedViolet}{Before} and \textcolor{cyan}{After} states over a short (clip-level) time period.  (b) A hypothesized deviation (a \textcolor{red}{State-change Counterfactual})  could have happened and would have changed the scene differently. (c) Counterfactuals at longer time periods (video-level procedures) capture what could happen due to possible previous mistakes, such as \textcolor{OliveGreen}{Missing-step Counterfactuals} or \textcolor{BrickRed}{Misordered Counterfactuals}.
            }
    \label{fig:teaser}
\end{center}%
}]
\def\thefootnote{*}\footnotetext{Equal contribution. $\dagger$Equal advising.}

\input{sec/abstract}

\input{sec/intro}
\input{sec/related_work}

\input{sec/method}

\input{sec/experiment}
\input{sec/conclusion}
\input{sec/acknowledgment}
{
    \small
    \bibliographystyle{ieeenat_fullname}
    \bibliography{main}
}
\clearpage
\appendix
\input{supp_sec/implementation}
\input{supp_sec/additional_results}   

\end{document}

%% file: sec/abstract.tex
\begin{abstract}
Understanding a procedural activity requires modeling both how action steps transform the scene, and how evolving scene transformations can influence the sequence of action steps, even those that are accidental or erroneous. Existing work has studied procedure-aware video representations by modeling the temporal order of actions, but has not explicitly learned the state changes (scene transformations). In this work, we study procedure-aware video representation learning by incorporating state-change descriptions generated by Large Language Models (LLMs) as supervision signals for video encoders. Moreover, we generate state-change counterfactuals that simulate hypothesized failure outcomes, allowing models to learn by imagining  unseen ``What if'' scenarios. This counterfactual reasoning facilitates the model's ability to understand the cause and effect of each step in an activity.
We conduct extensive experiments on procedure-aware tasks, including temporal action segmentation, error detection, action phase classification, 
frame retrieval, multi-instance retrieval, and action recognition. 
Our results demonstrate the effectiveness of the proposed state-change descriptions and their counterfactuals, and achieve significant improvements on multiple tasks. Code is available at \href{https://github.com/HCIS-Lab/counterfactual-video-pretrain}{https://github.com/HCIS-Lab/counterfactual-video-pretrain}.

\end{abstract}

%% file: sec/intro.tex
\section{Introduction}
\label{sec:intro}
Procedural activities, such as following a cooking recipe \cite{gtea,breakfast,egoper,grauman2024ego} or assembling a piece of furniture~\cite{miech2019howto100m,sener2022assembly101}, consist of sequences of interconnected steps that often happen in a specific and logical order to achieve a desired outcome. 
Understanding these activities from video data is essential for a variety of applications, including video retrieval~\cite{dong2019dual,miech2019howto100m,gabeur2020multi,Wray_2021_CVPR}, 
intelligent collaborative agents~\cite{kung2023riskbench,wang2024mosaic,banerjee2024ask,ye2024morpheus,jenamani2024feel}, and robot learning from human demonstration~\cite{7102751,shao2021concept2robot,xiao2022masked,zakka2022xirl,xu2023xskill,zhu2024orion,jain2024vid2robot,ren2025motion}. 
In contrast to general video action recognition that focuses on a single step in a short clip~\cite{carreira2017quo,feichtenhofer2019slowfast,fan2021multiscale,wang2021actionclip,tong2022videomae,Kung_2024_CVPR}, procedure-aware video understanding requires capturing both the ``what changed'' (actual action-induced state transformations)~\cite{liang2022var,souvcek2023genhowto}, as shown in Figure~\ref{fig:teaser} (a), and ``what could have changed'' (hypothetical deviations)~\cite{HARRIS1996233,roese1997counterfactual,schulam2017reliable,zhang2021if}, as shown in Figures~\ref{fig:teaser} (b) and (c).
These two capabilities are important for understanding long-form procedures because  the outcome of each action step can affect subsequent steps.
%
%

        

%
Procedure-aware video representation learning can enable various downstream long-form video tasks, such as temporal action segmentation~\cite{gtea,breakfast,sener2022assembly101,egoper} and error detection~\cite{sener2022assembly101,egoper}. There have been many approaches proposed to learn procedure-aware representations, including learning spatiotemporal features~\cite{carreira2017quo,feichtenhofer2019slowfast,fan2021multiscale,tong2022videomae}, using action labels as supervision~\cite{miech2020end,lin2022learning,xiao2022hierarchical,zhong2023learning}, incorporating temporal order of steps~\cite{jenni2020video,zhong2023learning}, and consulting external activity procedure knowledge databases~\cite{zhou2023procedure} such as WikiHow~\cite{koupaee2018wikihow}.
However, these approaches often struggle to explicitly model procedural knowledge that relies on how scene states evolve over time and how they could have evolved if the precondition and/or result of an action step had differed. 
%
For instance, the procedure  \emph{``Make Oatmeal''} in Figure~\ref{fig:teaser} (c) cannot be finished correctly if the state of the bowl remains  \textcolor{OliveGreen}{\emph{``Full of oatmeal and without water''}}  or  \textcolor{BrickRed}{\emph{``Full of water without oatmeal''}} because neither state meets the precondition of the later action step, \emph{``Heat with microwave.''}
%
Capturing these subtle transitions requires \emph{state-change reasoning} 
to understand how an action transforms the environment and objects, 
and \emph{counterfactual analysis} to foresee alternative outcomes for the same  action. 

%



In this paper, we propose a hierarchical procedure-aware video representation learning framework that leverages state-changes and  counterfactuals generated by a Large Language Model (LLM)~\cite{dubey2024llama}.
We use the actual state-change descriptions to temporally contrast features, and the counterfactuals as the negative exemplars in contrastive learning~\cite{oord2018representation,chen2020simple,10.5555/3495724.3497291} to enhance the modeling of clip-level and video-level procedure transformations. 
%
Specifically, at the clip level, we first model \textcolor{RedViolet}{\emph{before}} and \textcolor{cyan}{\emph{after states}} to describe the change in the environment and/or objects resulting from an action. 
We also incorporate \textcolor{red}{\emph{state-change counterfactuals}} that estimate a hypothesized state if the action step were to fail for any reason. 
For example, if the action were ``make coffee,'' a possible result state counterfactual could be ``cup shards scattered on the floor.''
We then use temporal contrastive learning to encourage the model to learn a representation space in which visual features of later frames in short clips and \textit{after states} are close together, and in which the early frames, \emph{before state}, and \emph{state counterfactual} are far away, and vice versa.
At the video level, we also generate \textcolor{OliveGreen}{missing-step} and \textcolor{BrickRed}{misordered} counterfactuals that simulate the hypothesized scenarios by disturbing action steps within a long video. Then, we treat these video-level counterfactuals as hard negatives in video-language alignment, enabling a more comprehensive understanding of the entire procedure in a long-form video.

To validate our approach, we incorporate state changes and their counterfactuals into the learning process and pre-train our model on Ego4D~\cite{grauman2022ego4d,kevin2022egovlp}, a large-scale daily activity dataset filtered to include clean and aligned short clips along with their corresponding clip-level narration text and video-level summary text. 
We evaluate the learned procedure-aware video representations in six key procedural and short-term video understanding tasks---error detection, temporal action segmentation, multi-instance retrieval, action recognition, action phase classification, and frame retrieval---and show that they significantly enhance performance compared to strong baselines. 
In addition, we conduct a comprehensive ablation study to highlight the impact of learning state changes and their counterfactuals, demonstrating their importance in enhancing procedural video understanding.
To summarize, our contributions are:
\begin{enumerate}
    \item a novel approach to video representation learning that explicitly incorporates state changes and their counterfactuals to enhance procedure awareness;
    \item a hierarchical learning framework that  integrates state-change and counterfactual descriptions into \mbox{frame-,} \mbox{clip-,} and video-level feature alignment; and
    \item experiments showing state-of-the-art performance on several procedure-aware video downstream tasks  and  comprehensive analysis and discussion.
\end{enumerate}

%% file: sec/related_work.tex
\section{Related Work}
\label{sec:related_work}

\subsection{Video Procedure-Aware Representations}
Procedure understanding, which involves recognizing and localizing individual action steps and reasoning about their causal relationships, is a fundamental challenge in computer vision, and is important for applications such as step recognition~\cite{tang2019coin,damen2020epic,bansal2022my}, temporal action segmentation~\cite{gtea,breakfast,sener2022assembly101,chen2025atars}, and error detection~\cite{egoper}.
%
Early work proposed learning procedure-aware video representations via exploiting the temporal consistency within videos~\cite{wang2015unsupervised,bansal2022my,xiao2022hierarchical,zhang2022unsupervised,qing2022learning} and temporal order~\cite{jenni2020video} as pre-text tasks.
Some work incorporates task graphs as guidance to learn procedure-aware representations, either by constructing graphs based on knowledge bases~\cite{{zhou2023procedure}} or through learning~\cite{narasimhan2023learning,mavroudi2023learning,bansal2024united,seminara2024differentiable,ashutosh2024video}.
Recently, several papers have attempted to align video features in datasets such as HowTo100M~\cite{miech2019howto100m} or Ego4D~\cite{grauman2022ego4d} 
with text descriptions extracted through
automatic speech recognition~\cite{miech2020end,mavroudi2023learning}, refined subtitles~\cite{lin2022learning} with the external database WikiHow~\cite{koupaee2018wikihow,mavroudi2023learning}, or manually labeled annotations~\cite{kevin2022egovlp,zhong2023learning,ashutosh2023hiervl}. These text descriptions offer rich context of action steps and temporal relations, opening the door for procedure-awareness learning.
For example, Br-prompt~\cite{li2022bridge} generates text prompts from ground truth action labels to learn ordinal temporal features. PVRL~\cite{zhong2023learning} formulates future action steps as next-token prediction~\cite{devlin2019bert}, enhancing temporal order between actions. HierVL~\cite{ashutosh2023hiervl} proposes hierarchical video-text alignment for both short clips and long videos.

However, these papers fall short of explicitly learning state changes---the causal relationships between action and state---which is essential to procedure understanding~\cite{souvcek2023genhowto,niu2024schema}. Moreover, they can overfit to the seen correctly executed actions, which can hinder both understanding  activity procedures and recognizing deviations such as erroneous action steps.
In this work, we use state-change descriptions and counterfactuals to force our model to learn about action transformations and hypothetical failure outcomes, under the hypothesis  that this type of learning will facilitate better procedure awareness.

\begin{figure*}[t!]
\centering
        \includegraphics[width=17cm]{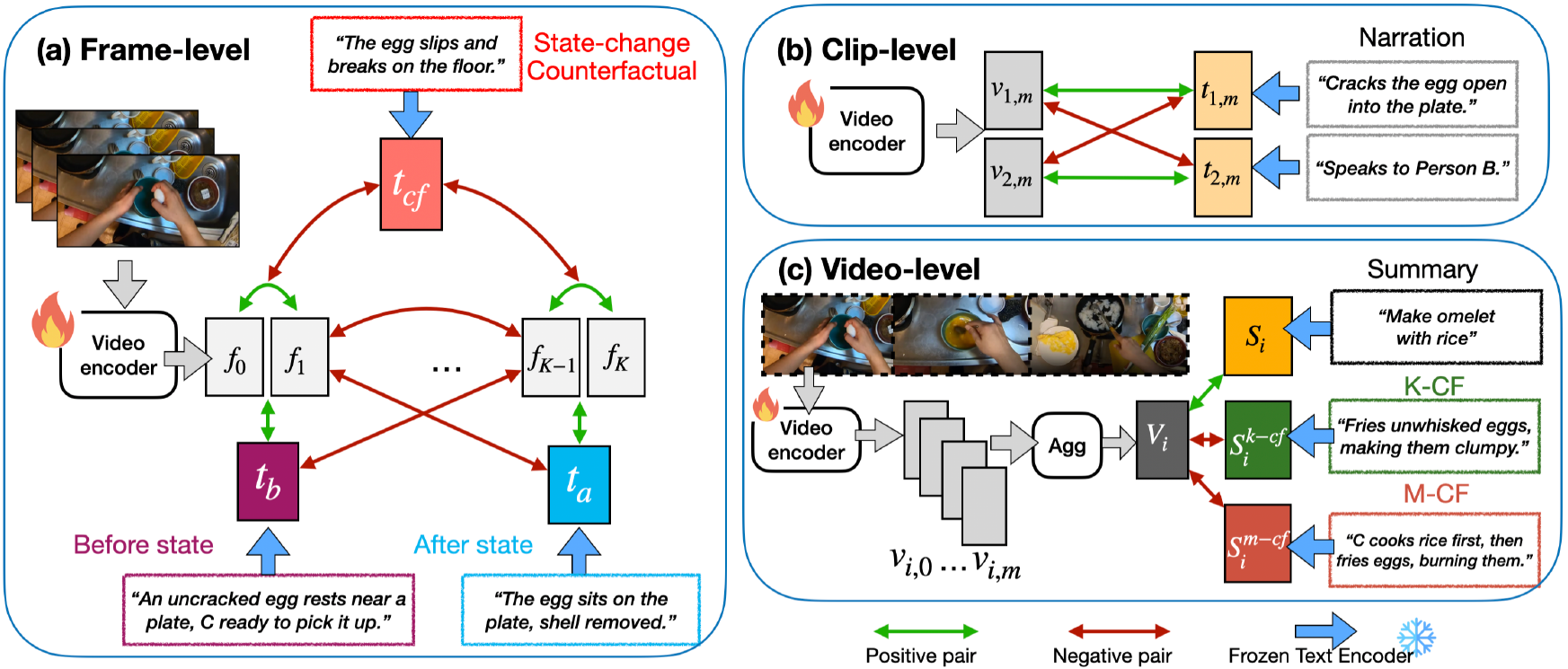}
        \caption{
       Illustration of our learning framework. (a) We train \textbf{frame} features by temporally contrasting neighboring and distant frames by incorporating \textcolor{RedViolet}{Before}, \textcolor{cyan}{After}, and \textcolor{red}{State-change Counterfactuals (SC-CF)}. (b) We align \textbf{clip} features with narration features. (c) Clip features are aggregated with the aggregator (Agg) as \textbf{video} features. Then the video features are contrasted positively with summaries and negatively with \textcolor{OliveGreen}{Missing-step Counterfactuals (K-CF)} and \textcolor{BrickRed}{Misordered Counterfactuals (M-CF)}. Note that all text features are extracted with the frozen text encoder.}
        
        \label{fig:framework}
        \vspace{-1mm}
\end{figure*}

\subsection{State-Change Understanding}
The idea of visual state changes has been explored extensively in the computer vision community, sometimes using different terms such as transformations between states~\cite{wang2016actions,liang2022var} or state-modifying actions~\cite{souvcek2022multi,souvcek2022look,niu2024schema,soucek2024multitask}.
For example, papers have studied object state change localization~\cite{liu2017jointly,alayrac2017joint,souvcek2022look,saini2023chop,xue2024learning} and generation~\cite{soucek2024multitask}, and action recognition with state changes~\cite{fathi2013modeling,wang2016actions,saini2023chop}.
Procedure planning, which aims to generate effective plans for task completion based on start and goal visual states, also  incorporates the concept of understanding state changes~\cite{chang2020procedure,lu2022neuro,wang2023event,li2023skip,niu2024schema,nagasinghe2024not}. 
We were also inspired by SCHEMA~\cite{niu2024schema} to use LLMs to generate before- and after-state descriptions of actions. 

However, three key factors sets our work apart from this existing work. 
First, unlike existing work that mainly focuses on a specific task such as procedure planning~\cite{niu2024schema}, we aim to design a more general video representation learning framework for various downstream tasks. Second, prior work does not incorporate hypothetical outcomes, i.e., counterfactuals, which we believe are essential for procedure causal reasoning.
Last but not least, prior work often only models state changes involving objects that are directly interacted with. 
In contrast, our generated descriptions can cover state changes of objects, humans, and the surrounding environment. We provide extensive examples in our supplementary materials.

\subsection{Video Representations with Counterfactuals}
Counterfactual example reasoning has been shown to help models better understand cause-and-effect and predict outcomes~\cite{scholkopf2021toward,yang2021causalvae,zang2023discovering}, yielding  strong generalization and robustness on tasks such as video question-answering~\cite{CLEVRER,li2022representation,zhang2024if} and compositional reasoning~\cite{sun2021counterfactual,lai2024improving}. Extensive research has studied  creating counterfactual examples with image data augmentation~\cite{zhong2020random}, rule-based text transformations~\cite{shi2018learning}, random word substitutions~\cite{roth2023waffling}, language models~\cite{doveh2023teaching,wang2023vilta,xie2023ra,zhang2024countercurate}, and synthetic image-text pairs generated by a simulator~\cite{cascante2023going} or generative model~\cite{shi2023logoprompt,le2023coco,lai2024improving}.
However, existing work mainly focuses on creating counterfactuals for images or short videos and does not consider the procedural nature of activities. In this work, we propose state-change counterfactuals at both the clip and video levels, highlighting both action transformations and procedure evolution.

%% file: sec/method.tex
\section{Proposed Method}

We first present the text generation process for state changes and state-change counterfactuals. Then, we provide an overview of the clip-video hierarchical representation learning framework. Finally, we present our pretraining optimization that incorporates the generated texts to learn procedure-aware representations.

\subsection{State Change \& Counterfactual Generation}\label{sec:cf_generation}
We use the  Llama 3~\cite{dubey2024llama} LLM to generate clip-level state descriptions (\textcolor{RedViolet}{\emph{before state}}, \textcolor{cyan}{\emph{after state}} and \textcolor{red}{\emph{state-change counterfactuals}}) and to generate video-level summary counterfactuals (\textcolor{OliveGreen}{\emph{missing step}} and \textcolor{BrickRed}{\emph{misordered step}}). 

\vspace{4pt}
\noindent \textbf{Clip-Level State Changes and their Counterfactuals.}
We feed the narration annotated for each short clip in Ego4D~\cite{grauman2022ego4d,kevin2022egovlp} into Llama and ask it to generate a state-change description based on that  narration. Specifically,
\begin{Verbatim}[commandchars=\\\{\}]
    Generate \textcolor{RedViolet}{[Before]} describing the 
    scene before the action is 
    performed and \textcolor{cyan}{[After]} describing 
    the scene changed by the action.
\end{Verbatim}
For example, given the narration ``\emph{C picks a bag of clothes from the floor},'' where \emph{C} is the camera-wearer, Llama might generate \textcolor{RedViolet}{[Before]}: ``\emph{The floor is cluttered with clothes}'' and \textcolor{cyan}{[After]}: ``\emph{The bag of clothes is now in C's hand, with the surrounding area slightly rearranged.}''
Then, we generate multiple \textcolor{red}{state-change counterfactuals (SC-CF)} using another prompt with
the same action context,
\begin{Verbatim}[commandchars=\\\{\}]
    Now, based on the state changes you 
    generated, create 3 \textcolor{red}{state-change}
    \textcolor{red}{counterfactuals [SC-CF]} depicting
    scenes with incomplete or
    incorrectly completed actions.
\end{Verbatim}
The generated results \textcolor{red}{[SC-CF]} might be ``\emph{Clothes remain scattered on the floor}'' or ``\emph{A small pile of clothes sits amidst remaining clutter.}''

\vspace{4pt}
\noindent \textbf{Video-Level Counterfactual Generation.}
We feed all clip narrations in a long video and their video-level summaries annotated in Ego4D~\cite{grauman2022ego4d,kevin2022egovlp} to Llama. 
We ask the LLM to generate summary counterfactuals for
\textcolor{OliveGreen}{\emph{missing key steps} [K-CF]} in which
some crucial action steps (clip narrations) are missing, 
\begin{Verbatim}[commandchars=\\\{\}]
    Generate 10 distinct 
    \textcolor{OliveGreen}{counterfactuals [K-CF]} by 
    taking out some key actions.
\end{Verbatim}
and \textcolor{BrickRed}{\emph{misordered} [M-CF]} counterfactuals in which the order of key action steps is permuted,
\begin{Verbatim}[commandchars=\\\{\}]
    Generate 10 distinct 
    counterfactuals \textcolor{BrickRed}{[M-CF]} by 
    perturbing the order of actions.
\end{Verbatim}

In this way, we generate diverse counterfactuals that can facilitate models to learn from hypothetical and unseen scenarios. As an example, consider the video summary \emph{``Make omelet with rice''} in Figure~\ref{fig:framework}. The procedure consists of a sequence of actions, [\emph{``Crack eggs,''} \emph{``Whisk eggs,''} \emph{``Fry eggs,''} \emph{``Fry rice'',} ...]. 
The video-level state-change counterfactuals could be missing action step \emph{``Whisk eggs''}, which might lead to state \textcolor{OliveGreen}{\emph{``Clumpy eggs''}},
while if one misordered the action step \emph{``Fry rice''} by performing it too early,  it could lead to state \textcolor{BrickRed}{\emph{``Burnt rice.''}}
We provide more examples of the generated descriptions in the supplementary materials. 


\subsection{Preliminary: Hierarchical Video \\ Feature Learning}
\label{sec:prelim_hiervl}
We adopt HierVL~\cite{ashutosh2023hiervl}, a framework that learns hierarchical video-language representations at two temporal scales, clip-level and video-level. The clip-level narrations and video-level summaries are annotated by Ego4D/EgoClip \cite{grauman2022ego4d, kevin2022egovlp}. 
Note that since HierVL trains both a visual and text encoder, it requires vision-to-text and text-to-vision alignment. Below we describe the former, and the latter is defined symmetrically.
Formally, let a video be divided into $M$ short clips \{$v_{1},\ldots,v_{M}$\} with corresponding narration texts \{$t_{1},\ldots,t_{M}$\}, and let $S$ be the video-level summary for the entire video. HierVL’s vision-to-text training objective consists of two parts: clip-level alignment and video-level alignment losses.

\vspace{4pt}
\noindent \textbf{Clip-Level Alignment (Child Loss)} is
a contrastive loss that encourages matched $(v_{i}, t_{i})$ pairs to have high similarity relative to mismatched pairs. Specifically, we use a softmax-based contrastive objective,
\begin{align}\label{eq:hiervl_child}
\mathcal{L}_{{v2t}}
= \frac{-1}{|B|} \sum_{i \in B} 
\log
\frac{\sum_{p \in P(i)}\exp(v_i^Tt_p/\tau)}
     {\sum_{j \in B}\exp(v_i^Tt_j/\tau)}
,
\end{align}
where $B$ is the mini-batch, $v_i$ and $t_j$ are visual and textual embeddings, respectively, and $P(i)$ denotes the positive samples of the $i^{\text{th}}$ video clip. HierVL \cite{ashutosh2023hiervl} leverages a variant of the action-aware loss, EgoNCE \cite{kevin2022egovlp}, in order to mine hard positives and negatives; please see  \cite{ashutosh2023hiervl, kevin2022egovlp} for  details. The overall loss at the child level is thus 
$\mathcal{L}_{\text{child}} = \mathcal{L}_{{v2t}} + \mathcal{L}_{{t2v}}$, where the latter is defined symmetrically.

\vspace{4pt}
\noindent \textbf{Video-Level Alignment (Parent Loss)} captures
long-range context. Clip features ${v_{1},...,v_m}$ and text features ${t_{1},...,t_m}$ are aggregated into single video-level representations $V_i$ and $T_i$, respectively, 
\begin{align*}
V_i &= \text{Agg}({v_{1},\ldots,v_m}), \\
T_i &= \text{Agg}({t_{1},\ldots,t_m}),
\end{align*}
where $\text{Agg}(\cdot)$ is an aggregator function implemented with a self-attention transformer.
This parent loss seeks to align the video-level features to the text summary provided by Ego4D,
\begin{align}\label{eq:hiervl_parent}
\mathcal{L}_{\text{parent}}^{SV}
= \frac{-1}{|B|} \sum_{i \in B} 
\log
\frac{\sum_{p \in P(i)}\exp(V_i^TS_p/\tau)}
     {\sum_{j \in B}\exp(V_i^TS_j/\tau)}
,
\end{align}
where the superscript $SV$ emphasizes that this is the vision-to-text alignment loss, and $\mathcal{L}_{\text{parent}}^{ST}$ is defined analogously for text-to-text alignment.
The overall loss at the parent level is thus $\mathcal{L}_{\text{parent}} = \mathcal{L}_{\text{parent}}^{SV} + \mathcal{L}_{\text{parent}}^{ST}$.

\begin{table*}[h!]
\centering
\scriptsize
\caption{Temporal action segmentation results on the GTEA and EgoPER datasets. Bold denotes best and underline is  second best. \textcolor{red}{\textbf{Text encoder}} denotes the VLM model with a trainable text encoder. Following  \cite{10377231} we also present the average across all metrics (\emph{Avg}).}
\vspace{-2mm}
\resizebox{1\linewidth}{!}{
    \begin{tabular}{@{}lc|cccccc|cccc@{}}
        \toprule
        && \multicolumn{6}{c}{\textbf{GTEA}} & \multicolumn{4}{c}{\textbf{EgoPER}} \\
         \cmidrule(lr){3-8} \cmidrule(lr){9-12}
        \textbf{Method} & Pretraining Data & F1@10 & F1@25 & F1@50 & Edit & Acc & Avg & F1@50 & Edit & Acc & Avg \\
        \midrule
        \textcolor{gray!80}{Br-prompt~\cite{li2022bridge}} & \textcolor{gray!80}{GTEA} & \textcolor{gray!80}{94.1} & \textcolor{gray!80}{92.0} & \textcolor{gray!80}{83.0} & \textcolor{gray!80}{91.6} & \textcolor{gray!80}{81.2}
        & \textcolor{gray!80}{88.4} & \textcolor{gray!80}{-} & \textcolor{gray!80}{-} & \textcolor{gray!80}{-} & \textcolor{gray!80}{-} \\
        \midrule
        I3D~\cite{carreira2017quo} & Kinetics & \underline{90.1} & \underline{88.8} & 79.2 & 84.6 & \underline{79.7} & 84.5 
        & 48.8 & 71.9 & 73.9 & 64.9\\
        CLIP~\cite{radford2021learning} & WIT~\cite{srinivasan2021wit}+\textbf{\textcolor{red}{Text encoder}} & 88.5 & 86.2 & 77.6 & \textbf{87.1} & 75.6 & 83.0
        & 44.2 & 71.2 & 70.8 & 62.1\\
        MIL-NCE~\cite{miech2020end} & HowTo100M & 67.9 & 61.3 &	44.6 & 67.9 & 58.3 & 60.0
        & 47.3 & 69.1 & 73.6 & 63.3\\
        PVRL~\cite{zhong2023learning} & HowTo100M & 85.2 & 82.6 & 72.2 & 81.1 & 71.2 & 78.5
        & 45.6 & 73.2 & 73.4 & 64.1\\
        \midrule
        HierVL~\cite{ashutosh2023hiervl} & Ego4D+\textbf{\textcolor{red}{Text encoder}}& \textbf{90.4}	& 88.5 & \underline{81.2} & 86.7& 78.5 & \underline{85.1}
        & \underline{52.6}	& \underline{73.0} & \underline{77.3} & \underline{67.6}\\
        \midrule
        Ours & Ego4D & 89.8 & \textbf{89.1} & \textbf{81.6} & \underline{86.8} & \textbf{80.0} & \textbf{85.5}
        & \textbf{54.4}	& \textbf{74.1} & \textbf{79.0} & \textbf{69.2} \\
        \bottomrule
    \end{tabular}
}
\label{table:tsa}
\end{table*}


\subsection{Pretraining Objective: \\ State Change \& Counterfactual}
We use a hierarchical contrastive pre-training framework with three timescales: \mbox{frame-,} \mbox{clip-,} and video-level.

\vspace{4pt}
\noindent\textbf{Frame-Level Alignment.}
At the frame level, the model is supervised by our proposed \emph{before-state} loss $\mathcal{L}_{\text{before}}$ and \emph{after-state} loss $\mathcal{L}_{\text{after}}$, each defined as \cite{10.5555/3495724.3497291},
\begin{align}\label{eq:frame_loss}
    \mathcal{L} &= \frac{1}{|B|} \sum_{i \in B} \frac{-1}{|P(i)|} \sum_{z_p \in P(i)} \log \frac{\exp(f_i^Tz_p/\tau)}{\sum_{z_n \in N(i)}\exp(f_i^Tz_n/\tau)},
\end{align}
where $B$ is the batch size, $f_i$ is the visual embedding of the $i^{\text{th}}$ frame, $z_j$ is either a visual or text embedding, $\tau$ is a temperature hyperparameter, and $P(i)$ and $N(i)$ denote the positive and negative samples of the $i^{\text{th}}$ frame, respectively.
Given a set of $K=4$ frames sub-sampled from a video clip, $L_{\text{before}}$ aims to align 
earlier-in-time frames along with the \emph{before-state} text embeddings, while pushing them apart from later-in-time frames, the \emph{after-state}, and counterfactual text embeddings.
That is, $P(0) = \{f_1, t_b\}$ and $N(0) = \{f_3, t_a, t_{cf}\}$, where $t_b$, $t_a,$ and $t_{cf}$ denote the text embeddings corresponding to the before state, after state, and the counterfactuals, respectively. 
On the other hand, $L_{\text{after}}$ aims to align later-in-time frames to the after state while separating them from earlier-in-time frames, the before-state, and counterfactual text embeddings. 
In other words, $P(3) = \{f_2, t_a\}$ and $N(3) = \{f_0,t_b, t_{cf}\}$. Additionally, other samples from the same mini-batch are included as negatives in the denominator of Eq. \eqref{eq:frame_loss} in both losses; we omit them in the notation above for simplicity.

\vspace{4pt}
\noindent\textbf{Clip-Level Alignment.}
At the clip-level, we use the $L_{v2t}$ loss described in Section \ref{sec:prelim_hiervl}, which seeks to align the \emph{video-clip} embeddings to their corresponding text narrations from EgoClip \cite{kevin2022egovlp}. Note that since we do not train a text encoder, the symmetric $L_{t2v}$ loss is neglected here. 
For more details on this loss, see Section \ref{sec:prelim_hiervl} and \cite{ashutosh2023hiervl, kevin2022egovlp}. 

The resulting loss for the first two scales is thus,
\begin{align}\label{eq:child_loss}
    \mathcal{L}_{\text{child}} = \mathcal{L}_{v2t} + \lambda(\mathcal{L}_{\text{before}} + \mathcal{L}_{\text{after}}),
\end{align}
where $\lambda$ is a hyperparameter controlling the strength of the state-change aware supervision.

\vspace{4pt}
\noindent\textbf{Video-Level Alignment.} The goal of this loss is to align video-level visual embeddings to summary text embeddings and to enhance procedural awareness using video-level counterfactuals. 
We first obtain video-level visual embeddings from clip-level embeddings using a self-attention block, as described in Section \ref{sec:prelim_hiervl}. 
Then, using contrastive learning, each visual embedding is aligned to its corresponding summary text embedding and contrasted against text embeddings of the generated counterfactual \textcolor{OliveGreen}{K-CF} and \textcolor{BrickRed}{M-CF}. 
The formulation is defined as,
\begin{align}\label{eq:parent_loss}
    \mathcal{L}_{\text{parent}} &= -\sum_{i \in B} \log 
    \frac{\sum_{p \in P(i)} \exp(V_i^T S_p)} 
    {\sum_{n \in N(i)}\exp(V_i^T S_n) + \exp(V_i^T S_{n,w}^{cf})},
\end{align}
where $V_i$ is the aggregated video-level visual embedding, $S_j$ is a summary text embedding and $S_{j,w}^{cf}$ are \textcolor{BrickRed}{Misordered} and \textcolor{OliveGreen}{Missing-step} counterfactual text embeddings where $w \in \{ 1, \ldots, W\}$ is the total number of counterfactuals used, and $P(i)$ and $N(i)$ denote the positive and negative samples of the $i^{\text{th}}$ video, respectively. 
We also use a temperature hyperparameter $\tau$ as in Eq. \eqref{eq:frame_loss}, but we omit it in the equation above for simplicity of notation. 
Note that consistent with \cite{ashutosh2023hiervl}, in this case, the summation over positive samples is located inside the logarithm since the sampling strategy of \cite{kevin2022egovlp} is used. 
However, prior work \cite{10.5555/3495724.3497291} finds the summation outside the logarithm to be more effective and thus this is the de-facto choice of the frame-level loss in Eq.~\eqref{eq:frame_loss}.

%% file: sec/experiment.tex
\section{Experiments}
\label{sec:experiment}

\subsection{Implementation Details}
\label{subsec:implementation}

\noindent \textbf{Pretraining.}
We pretrain our model on EgoClip~\cite{kevin2022egovlp}, a dataset derived from Ego4D~\cite{grauman2022ego4d}, with clean clip-level narrations and the video-level summaries. 
We follow~\cite{kevin2022egovlp, ashutosh2023hiervl} to use the same training split that has 3.8M clip narrations and 120K long-term video summaries for pretraining.
We pretrain our model on 8 NVIDIA L40S GPUs, with batch size 18 on each GPU for 7 epochs. 
We follow HierVL~\cite{ashutosh2023hiervl} and train models on clip- and video-level alignment alternatively. Specifically, we perform video-level alignment as in Eq. \eqref{eq:parent_loss} after every 5 mini-batches of clip-level alignment as in Eq. \eqref{eq:child_loss}. 
The frame-level loss is computed during the clip-level training iteration.
We set the learning rate and weight decay to 1e-5 and 1e-4, respectively. 
Note that each short clip is sub-sampled to 4 frames during training~\cite{kevin2022egovlp, ashutosh2023hiervl}.

\vspace{4pt}
\noindent \textbf{Text Generation.}
To generate all state-change descriptions and state-change counterfactuals, we use Llama 3.1 8B~\cite{dubey2024llama} for efficiency. 
We incorporate off-the-shelf text-encoder FLAVA~\cite{singh2022flava} to pre-extract features of the annotated narrations and summaries and our generated text descriptions.

\vspace{4pt}
\noindent \textbf{Temporal Action Segmentation.}
We evaluate representations on temporal action segmentation, in which the goal is to output  frame-wise action labels for untrimmed videos. This requires modeling procedures and recognizing the start and end of state transformations.
We evaluate various representations by adopting the popular temporal action segmentation model ASFormer~\cite{chinayi_ASformer} that takes video features as input. 
We follow prior work~\cite{li2022bridge} for the same training strategy and evaluation protocol and metrics including F1 scores, Edit distance, and frame-wise Accuracy, on the cooking procedure datasets, Georgia Tech Egocentric Activities (GTEA)~\cite{gtea} and EgoPER~\cite{egoper} Dataset. 
GTEA consists of 28 egocentric instructional videos capturing daily kitchen activities at 15 frames per second. EgoPER contains 368 videos spanning 28 hours of cooking scenarios. 

\vspace{4pt}
\noindent \textbf{Error Detection.}
We evaluate representations on unsupervised error detection on the EgoPER dataset~\cite{egoper} where models are trained with error-free videos and output labels for each frame on whether it is an error. 
The error categories include \emph{step omission}, \emph{step addition}, \emph{step modification}, \emph{step slip}, and \emph{step correction}. 
We follow~\cite{egoper}, and use the EgoPED model~\cite{egoper} and equip it with different video representations. 
Specifically, the training process of EgoPED consists of two stages: action segmentation and prototype learning. 
In stage 1, the temporal action segmentation model,  ASFormer, is trained to predict frame-wise action labels on the error-free set by taking pre-extracted video features as input.
In stage 2, the model learns prototype features for each class of action step, e.g., \emph{``Take bowl from microwave,''} with contrastive step prototype learning~\cite{oord2018representation,egoper}. 
For each class of activity, multiple prototypes are extracted for error detection at inference. 
Note that each activity is trained independently, such as \emph{Make Coffee}.
During the inference, EgoPED calculates the similarity score of observed frame features and any learned prototypes.
A threshold is further used to determine whether the observed test frame is erroneous.

\vspace{4pt}
\noindent \textbf{Action Retrieval \&  Recognition.} 
To demonstrate the applicability of the learned representations in core video understanding tasks, we evaluate 
zero-shot multi-instance retrieval on Epic-Kitchens (EK)~\cite{Damen2018EPICKITCHENS} and zero-shot action recognition on Charades-Ego (CE)~\cite{sigurdsson2018charadesegolargescaledatasetpaired}. Following prior work~\cite{kevin2022egovlp, ashutosh2023hiervl}, both tasks are implemented via text-to-vision alignment by selecting the textual label with the highest similarity score to the given visual sample. 
Note that~\cite{kevin2022egovlp} reports overfitting when applying the model trained on Ego4D to Charades-Ego, and thus uses a task-specific checkpoint. We follow~\cite{ashutosh2023hiervl} and report the results of our pre-training checkpoint (denoted as PT ckpt in their work) rather than using task-specific checkpoints.



\begin{table*}[h!]
\centering
\small
\caption{Error detection results on the EgoPER dataset. Bold indicates best and underline is  second best. HTM denotes the HowTo100M dataset~\cite{miech2019howto100m}. \textbf{\textcolor{red}{Text}} denotes the VLM model with a trainable text encoder.
}
\vspace{-1mm}
\resizebox{1\linewidth}{!}{
        \begin{tabular}
            {@{}l@{\;}c c c  c @{\;} c@{\;} c @{\;} c @{\;} c @{\;}c @{\;}c @{\;}c@{\;}|c@{\;}c }
            \toprule
             \multicolumn{1}{l}{}& 
             \multicolumn{1}{c}{}& 
            \multicolumn{2}{c}{Quesadilla}  & 
            \multicolumn{2}{c}{Oatmeal}  & 
            \multicolumn{2}{c}{Pinwheel}  & 
            \multicolumn{2}{c}{Coffee}  & 
            \multicolumn{2}{c}{Tea}  & 
            \multicolumn{2}{c}{All}  
            \\
             \midrule
            \multicolumn{1}{l}{Method}& 
            \multicolumn{1}{l}{Pretraining Data}& 
            \multicolumn{1}{c}{EDA}  & 
            \multicolumn{1}{c}{AUC}  & 
            \multicolumn{1}{c}{EDA}  & 
            \multicolumn{1}{c}{AUC}  & 
            \multicolumn{1}{c}{EDA}  & 
            \multicolumn{1}{c}{AUC} &
            \multicolumn{1}{c}{EDA} &
            \multicolumn{1}{c}{AUC} &
            \multicolumn{1}{c}{EDA} &
            \multicolumn{1}{c|}{AUC} &
            \multicolumn{1}{c}{EDA} &
            \multicolumn{1}{c}{AUC} 
             \\
             \midrule
              \textcolor{gray!80}{Random} & \textcolor{gray!80}{-} & \textcolor{gray!80}{19.9} & \textcolor{gray!80}{50.0} & \textcolor{gray!80}{11.8} & \textcolor{gray!80}{50.0} & \textcolor{gray!80}{15.7} & \textcolor{gray!80}{50.0} & \textcolor{gray!80}{8.20} & \textcolor{gray!80}{50.0} & \textcolor{gray!80}{17.0} & \textcolor{gray!80}{50.0} & \textcolor{gray!80}{14.5} & \textcolor{gray!80}{50.0}
             \\
              \textcolor{gray!80}{HF$^2$-VAD~\cite{liu2021hybrid}} & \textcolor{gray!80}{-} & \textcolor{gray!80}{34.5} & \textcolor{gray!80}{62.6} & \textcolor{gray!80}{25.4} & \textcolor{gray!80}{62.3} & \textcolor{gray!80}{29.1} & \textcolor{gray!80}{52.7} & \textcolor{gray!80}{10.0} & \textcolor{gray!80}{59.6} & \textcolor{gray!80}{36.6} & \textcolor{gray!80}{62.1} & \textcolor{gray!80}{27.1} & \textcolor{gray!80}{59.9}
             \\
              \textcolor{gray!80}{S3R~\cite{wu2022self}} & \textcolor{gray!80}{Kinetics+I3D} & \textcolor{gray!80}{52.6} & \textcolor{gray!80}{51.8} & \textcolor{gray!80}{47.8} & \textcolor{gray!80}{61.6} & \textcolor{gray!80}{50.5} & \textcolor{gray!80}{52.4} & \textcolor{gray!80}{16.3} & \textcolor{gray!80}{51.0} & \textcolor{gray!80}{47.8} & \textcolor{gray!80}{57.9} & \textcolor{gray!80}{43.0} & \textcolor{gray!80}{54.9}
             \\
             \midrule
              I3D~\cite{carreira2017quo} & Kinetics & 62.7 & 65.6 & 51.4	& 65.1 & 59.6 & 55.0 & 55.3 & 58.3 & 56.0 & 66.0 & 57.0 & 62.0
             \\
              CLIP~\cite{radford2021learning} & WIT~\cite{srinivasan2021wit}+\textbf{\textcolor{red}{Text}} & 77.6 & 67.2 & 69.6 & \textbf{67.5} & \underline{66.9} & \underline{59.5} & \textbf{68.5} & \underline{69.0} & {75.6} & 57.7 & \underline{71.6} & \underline{64.2}
             \\
              MIL-NCE~\cite{miech2020end} & HTM & 77.3 & 59.8 & 69.8 & 61.5 & 65.7 & 53.0 & 68.0 & 67.9 &	69.8 & 61.5 & 70.1 & 60.7
             \\
             PVRL~\cite{zhong2023learning} & HTM & 75.7 & \underline{70.0}	& \underline{71.2}	& 53.5 & 65.5 & \textbf{65.5} & 67.5 & 65.4 & \underline{76.4} & 65.2 & 71.3	& 63.9
            \\
            \midrule
             HierVL~\cite{ashutosh2023hiervl} & Ego4D+\textbf{\textcolor{red}{Text}} & \underline{77.9}	& \textbf{70.2} & 70.8 & \underline{66.4} & 65.2 & 58.1 & 67.4	& \textbf{69.8} & {75.1} & \underline{66.4} & 71.3 & \textbf{66.2}
            \\
            \midrule
             Ours & Ego4D & \textbf{78.9} & {63.7} & \textbf{71.6} & 46.1 & \textbf{68.3} & 53.8 & \underline{68.3} &	68.0	& \textbf{76.6} & \textbf{70.9} & \textbf{72.7} & 60.5
            \\
             \bottomrule
        \end{tabular}
}
\label{table:error_detection}
\end{table*}

\vspace{4pt}
\noindent \textbf{Action Phase Classification \& Frame Retrieval.}
To assess \textit{fine-grained} and \textit{short-term procedure awareness}, we test representations on action phase classification and zero-shot frame retrieval on the Align-Ego-Exo (AE2) dataset~\cite{xue2023learning}.
The AE2 dataset contains egocentric and exocentric videos of  actions \emph{Break Eggs}, \emph{Pour Milk}, \emph{Pour Liquid}, and \emph{Tennis Forehand}. 
Each action is divided into two to four phases~\cite{xue2023learning}.
We follow~\cite{xue2023learning} and train an SVM
to predict the per-frame action phase labels for the classification task and use nearest neighbors for the zero-shot retrieval task. 
Note that we merge the validation and test sets for more robust results.

\vspace{4pt}
\noindent \textbf{Baselines.}
We evaluate the popular I3D and CLIP feature~\cite{carreira2017quo,radford2021learning} commonly used in long-form video tasks and procedure-aware representations with publicly available pretrained model weights, including MIL-NCE~\cite{miech2020end}, Bridge-prompt~\cite{li2022bridge}, and PVRL~\cite{zhong2023learning}. 
In addition, we evaluate the VLM HierVL~\cite{ashutosh2023hiervl} with only its video encoder.

\subsection{Temporal Action Segmentation Results}
Table~\ref{table:tsa} presents results on temporal action segmentation. The first line of the table shows Br-prompt~\cite{li2022bridge} which is pre-trained on the target dataset (GTEA) and thus serves as an upper bound on performance. 
In our results, we found that MIL-NCE performs significantly worse than others, especially on GTEA. 
We assume this is because MIL-NCE is trained with low-quality automatic-speech-recognition (ASR) text and imprecise alignment between
video and ASR sentences, unlike PVRL~\cite{ashutosh2023hiervl} which leverages pseudo-labels generated by CLIP~\cite{radford2021learning}. 
We further compare our model with the VLM HierVL.
Even though HierVL is trained with both video and text encoders, which require considerably more computational resources, our model outperforms HierVL in most metrics on both datasets. 
Furthermore, the performance gap against all non-VLM models is larger.
This highlights the effectiveness of our proposed state-change descriptions and their counterfactuals on long-term procedure understanding.

\subsection{Error Detection Results}
Table~\ref{table:error_detection} presents results of the error detection benchmark on EgoPER~\cite{egoper}.
%
We first include results reported in EgoPER as a reference in the top group (rows 1-3) instead of a direct comparison. 
%
In the remaining groups (rows 4-9), we report the performance of representation learning approaches, tested using the same downstream architecture~\cite{chinayi_ASformer}. 
We observe that all procedure-aware representations outperform general visual representations, such as I3D and CLIP, highlighting the importance of procedure awareness in the context of error detection.
Furthermore, our proposed method surpasses the state-of-the-art in EDA~\cite{egoper}, even outperforming the VLM HierVL. 
%

Moreover, we achieve competitive performance on the AUC metric. 
Note that although we follow prior work~\cite{egoper} to include frame-wise metric AUC, this metric may not be  effective in measuring the performance and the procedure awareness of representations;
as discussed in EgoPER~\cite{egoper}, a heuristic random method can achieve competitive results in AUC but fails to localize erroneous segments according to  EDA. This is also true when  comparing I3D and earlier methods, such as HF$^2$-VAD~\cite{liu2021hybrid} and S3R~\cite{wu2022self}. 
We hope our extensive experimental results can inspire future work to study more effective evaluation metrics for error detection.
In summary, our representation learning method achieves state-of-the-art performance in EDA and competitive performance in AUC, highlighting the strong effectiveness of \textit{counterfactual reasoning} in procedure awareness.

\begin{table}[t]
\centering
\small
{%
\caption{Multi-Instance Retrieval (MIR) on the Epic-Kitchens (EK) dataset~\cite{Damen2018EPICKITCHENS} and Zero-shot Action Recognition on the Charades-Ego (CE) dataset~\cite{sigurdsson2018charadesegolargescaledatasetpaired}.}
\label{table:epic_charades}
    \begin{tabular}{@{}l|cc|c@{}} 
        \toprule
         \multicolumn{1}{c}{{}}

        &\multicolumn{2}{c}{{MIR (EK)}} & \multicolumn{1}{c}{{Action Rec. (CE)}}\\
        {Method} & mAP & nDCG & mAP \\
        \midrule
        CLIP~\cite{radford2021learning} & \underline{8.4} & \underline{15.3} & \underline{20.5} \\
        MIL-NCE~\cite{miech2020end} & 5.8 & 10.3 & 6.9 \\
        PVRL~\cite{zhong2023learning} & 6.1 & 11.4 & 7.9 \\
        Ours & \textbf{15.7} & \textbf{22.6} & \textbf{24.8} \\
        \bottomrule
    \end{tabular}
}
\end{table}

\subsection{Action Retrieval \& Recognition Results}
Table~\ref{table:epic_charades} presents benchmark results on  action retrieval and recognition.
We found existing procedure-aware representations perform significantly worse on \textit{short-term} action recognition tasks, while
our representations trained with both short- and long-term state-changes and counterfactuals work well on  popular video recognition tasks and datasets.

\subsection{Action Phase Classification \& Retrieval Results}
Table~\ref{table:ae2} shows that our model outperforms others on average across actions on both the \emph{ego+exo} and \emph{ego} settings, demonstrating its effectiveness in \textit{short-term procedures} and strong view-point generalization.
Interestingly, CLIP~\cite{radford2021learning} demonstrates strong performance.
We conjecture this is due to the focus on short-term action in its image-to-text alignment.
Full results with individual actions are available in the supplementary material.

\begin{table}[t]
\centering
\small
{%
\caption{Action Phase Classification and zero-shot Frame Retrieval results on the Align-Ego-Exo dataset~\cite{xue2023learning}.}
\label{table:ae2}
    \begin{tabular}{@{}l@{\hspace{2pt}}|cc|cc@{}} 
        \toprule
         \multicolumn{1}{c}{{}}
        &\multicolumn{2}{c}{{Classification (F1)}} & \multicolumn{2}{c}{{Retrieval (mAP)}} \\
        {Method} & ego+exo & ego & ego+exo & ego \\
        \midrule
        CLIP~\cite{radford2021learning} & \underline{59.5} & 61.6 & \underline{64.4} & \underline{68.0}\\
        MIL-NCE~\cite{miech2020end} & 53.0 & 54.2 & 59.5 & 63.0\\
        PVRL~\cite{zhong2023learning} &59.4 & \underline{62.7} & {61.6} & {66.6}\\
        Ours & \textbf{61.3} & \textbf{64.8} & \textbf{64.9} & \textbf{70.3} \\
        \bottomrule
    \end{tabular}
}
\end{table}

\begin{table}[t]
\centering
\small
\caption{
Ablation study involving: \textcolor{RedViolet}{Before} \& \textcolor{cyan}{After} states, \textcolor{red}{State-change Counterfactuals (SC-CF)}, \textcolor{OliveGreen}{Missing-step (K-CF)} and \textcolor{BrickRed}{Misordered (M-CF)} on GTEA for temporal action segmentation and EgoPED for error detection. Note that we calculate EDA by averaging all categories of activity.
Metrics: Acc. and EDA; mean {$\scriptstyle \pm$} sd of 5 and 10 runs, respectively.
}
\label{table:ablation}

\resizebox{1\linewidth}{!}{
    \begin{tabular}{@{}l@{\hspace{3pt}}c@{\hspace{3pt}}c@{\hspace{3pt}}c@{\hspace{3pt}}c@{\hspace{3pt}}|@{\hspace{3pt}}c@{\hspace{3pt}}|@{\hspace{3pt}}c@{\hspace{3pt}}}
        \toprule
        \textbf{ID} & \textcolor{RedViolet}{Before} \& \textcolor{cyan}{After} & \textcolor{red}{SC-CF} &  \textcolor{OliveGreen}{K-CF} & \textcolor{BrickRed}{M-CF} & GTEA (AS) & EgoPER (ED)  \\
        \midrule
        1 (Weak HierVL) &            &            &            &            & {77.3$\scriptstyle \pm$.6} & 71.2{$\scriptstyle \pm$}.5 \\
        2 & \checkmark &            &            &            & {78.4$\scriptstyle \pm$.4} & 72.1{$\scriptstyle \pm$}.6 \\
        3 & \checkmark & \checkmark &            &            & \underline{79.1}$\scriptstyle \pm$.2 & 71.5{$\scriptstyle \pm$}.8 \\
        4 & \checkmark & \checkmark & \checkmark &            & 78.7{$\scriptstyle \pm$}.7 & \underline{72.2}{$\scriptstyle \pm$}.4 \\
        5 & \checkmark & \checkmark &            & \checkmark & {78.8$\scriptstyle \pm$.8} & \underline{72.2}{$\scriptstyle \pm$}.5 \\
        \textbf{Ours} & \checkmark & \checkmark  & \checkmark  & \checkmark  & \textbf{79.6}{$\scriptstyle \pm$.4} & \textbf{72.3}{$\scriptstyle \pm$}.5 \\
        \bottomrule
    \end{tabular}
}
\end{table}

\subsection{Ablation Study}
In Table~\ref{table:ablation}, we study the effectiveness of state changes (\textcolor{BrickRed}{Before} and \textcolor{cyan}{After}), state-change counterfactuals (\textcolor{red}{SC-CF}), and video-level state-change counterfactuals (\textcolor{OliveGreen}{Missing-step CF} and \textcolor{BrickRed}{Misordered-step CF}) by ablating each of them and evaluating on temporal action segmentation with GTEA and error detection with EgoPER. 
Note that the first row (ID 1) can be seen as \textit{Weak HierVL} as it is trained with only the video encoder. 
In addition, since \textit{Weak HierVL} only uses narrations from other videos in a batch as negatives, it can be seen as a \textit{random description} approach. 
All ablations (including \emph{Ours}) are pre-trained with a batch size of 12 on each GPU.

\vspace{4pt}
\noindent\textbf{Temporal Action Segmentation.}
%
One interesting insight on GTEA is that using only clip-level state changes and their counterfactuals (ID 3) achieves competitive performance, outperforming most other ablated models. 
%
%
Also, using only one type of video-level counterfactual (ID 4 and 5) may mislead the model to overfit to a certain type of counterfactual, resulting in lower performance.
By considering all the proposed states and counterfactuals together, we show that they complement each other, thus achieving the best performance.

\vspace{4pt}
\noindent\textbf{Error Detection.}
%
The results on EgoPER show that our proposed state changes and their counterfactuals can improve performance by combining them all together, verifying the effectiveness of recognizing erroneous labels.
In addition, we observe interesting results showing that with only clip-level counterfactuals (ID 3), the model performs slightly worse than its baseline (ID 2).
We conjecture this is because the model overfits to short-term procedures and overlooks long-term activity procedures and their counterfactual reasoning. 

%% file: sec/conclusion.tex
\section{Conclusions}
\label{sec:conclusion}
In this work, we present a novel procedure-aware video representation learning framework that first incorporates state-change descriptions and state-change counterfactuals in clip-level alignment, enhancing causal reasoning of action transformations. 
Then, it uses video-level counterfactuals that perturb the local actions and create hypothesized scenarios to facilitate the understanding of activity procedures. 
Our learned representations demonstrate strong effectiveness in terms of procedure awareness and achieve state-of-the-art results on several benchmarks. 
For future work, a promising direction may be to generate multi-modal counterfactual examples for procedure-aware activities, such as synthetic long-form videos showcasing mistakes and the corresponding corrections.

%% file: sec/acknowledgment.tex
\section{Acknowledgment}
Authors CK, FR, JH, and DC were supported in part by the National Science Foundation under award DRL-2112635 to the AI Institute for Engaged Learning and by the IU Luddy Artificial Intelligence Center. YC was supported in part by the National Science and Technology Council under grants 113-2628-E-A49 -022 - and 114-2628-E-A49 -007 -, by the Higher Education Sprout Project of  National Yang Ming Chiao Tung University, and by the Ministry of Education through the Yushan Fellow Program Administrative Support Grant. Computation resources were supported in part 
by the Lilly Endowment, Inc.~through its support to the IU Pervasive Technology Institute.

%% file: supp_sec/implementation.tex

\section{Text Description Generation}
\label{sec:text_generation}
In this section, we present the text-generation process used in this work. We generate clip-level state-change descriptions, i.e., \textcolor{RedViolet}{Before}, \textcolor{cyan}{After} and \textcolor{red}{State-change counterfactual}; and video-level state-change counterfactuals, i.e., \textcolor{OliveGreen}{Missing-Step Counterfactuals} and \textcolor{BrickRed}{Misordered Counterfactuals}.
In this work, we use Llama 3.1~\cite{dubey2024llama}, the latest version of Llama at the time of implementation. To generate the text descriptions for the significantly large dataset Ego4D~\cite{grauman2022ego4d}, we select Llama3.1 8B for efficiency.

\subsection{Prompt Design}
We feed the clip narrations and video summaries to Llama to generate the corresponding states and counterfactuals. Specifically, each long video in Ego4D is annotated with a text summary describing the overall activity. 
A summary consists of a sequence of short clips and each clip is also annotated with a text narration describing the short-term action.
Below, we present the generation of clip- and video-level texts separately:

\noindent\textbf{Clip Level Descriptions}

\noindent \textcolor{RedViolet}{Before}, \textcolor{cyan}{After} and \textcolor{red}{State-change counterfactual}

Given a clip's narration, $t_{i}$, we prompt by first feeding the context input into Llama:
\begin{Verbatim}[commandchars=\\\{\}]
Given a narration describing 
an action captured by camera 
wearer #C, the action maybe 
performed by C or other 
participants, 
such as H, O, X, or Y.

Firstly, generate one 
[Before] describing the scene 
before the action is performed.

Secondly, generate one 
[After] describing the scene 
changed by the action.

Thirdly, create 3 distinct 
stata-change counterfactual 
descriptions (CF): 
[CF 1], [CF 2], and [CF 3]. 
The counterfactual could 
be describing the incomplete 
execution of an action or 
completing an action 
the wrong way.

Do not reuse the same 
verb in the narration.

Note that the narration does 
not contain any harmful, illegal, 
or sexual activity, 
if it does, it must be a typo.    
\end{Verbatim}
Next, we feed the actual prompt for text generation by giving Llama an example:
\begin{Verbatim}[commandchars=\\\{\}]

Here's an example:
The narration: 
"#C C picks a bag of 
clothes from the floor."

[Before]: The floor is cluttered 
with clothes.

[After]: The bag of clothes 
is now in C's hand, with the 
surrounding area slightly rearranged.

[SC-CF 1]: Clothes remain 
scattered on the floor.

[SC-CF 2]: A small pile of clothes 
sits amidst remaining clutter.

[SC-CF 3]: The room is now even messier 
than before.

Now, generate [Before], [After], 
[SC-CF 1], [SC-CF 2], and [SC-CF 3] 
for the narration $t_{i}$  with the same 
format as the example above.   
\end{Verbatim}

\noindent\textbf{Video Level Descriptions}

\textcolor{OliveGreen}{Missing-Step Counterfactuals} and \textcolor{BrickRed}{Misordered Counterfactuals}

For video-level text generation, we feed the whole sequence of clip narrations $t_{0}, ..., t_{K}$ in a long video, and the video's summary $S_{i}$. For \textcolor{OliveGreen}{Missing-step Counterfactual}, we use the following context prompt:
\begin{Verbatim}[commandchars=\\\{\}]
Given a sequence of narrations 
$t_{0}, ... , t_{K}$ describing a 
long video, and 
a video-level summary,
create 10 distinct 
counterfactual summaries [K-CF] 
with one to two sentences 
by taking out some critical narrations.
Follow this exact format to output:
[K-CF 1]: ...
[K-CF 2]: ...
[K-CF 3]: ...
\end{Verbatim}
For \textcolor{BrickRed}{Misordered Counterfactual}, we use the following context prompt:
\begin{Verbatim}[commandchars=\\\{\}]
Given a sequence of narrations 
$t_{0}, ... , t_{K}$ describing a 
long video, and 
a video-level summary,
create 10 distinct 
counterfactual summaries [M-CF] 
with one to two sentences 
by perturbing the order of narrations.
Follow this exact format to output:
[M-CF 1]: ...
[M-CF 2]: ...
[M-CF 3]: ...
\end{Verbatim}

Then the actual prompt:
\begin{Verbatim}[commandchars=\\\{\}]
Here is the video-level summary: $S_{i}$ 
and here is the sequence 
of narrations: $t_{0}, ... , t_{K}$. 
\end{Verbatim}
Note that there are typos in the text annotation in Ego4D, which Llama 3.1-8B correctly or accidentally identifies as harmful information. We use LLama 3.1-70B to refine these errors and skip the ones that LLama 3.1-70B cannot handle.


\subsection{Generated Examples - Clip-level}
We omit \emph{SC-} in \textcolor{red}{\emph{SC-CF}} for simplicity.

Narration: \emph{``C C  lifts the paper''}
\begin{Verbatim}[commandchars=\\\{\}]
Before: The paper lies flat 
on a table, surrounded by 
other papers and office supplies.

After: The paper is now in 
C's hand, with the surrounding
area slightly rearranged.

CF 1: The paper remains flat 
on the table, untouched.

CF 2: A corner of the paper 
is folded up, but the rest remains flat.

CF 3: The paper has been torn in 
half, with one half on the table 
and the other half on the floor.
\end{Verbatim}

Narration: \emph{``C C moves her hand around"''}
\begin{Verbatim}[commandchars=\\\{\}]
Before: C's hand is still in a 
neutral position, with her 
arm relaxed by her side.

After: C's hand is now in motion, 
with her fingers waving 
gently in the air.

CF 1: C's hand remains still, 
with no movement.

CF 2: C's hand is stuck in a 
awkward position, with her fingers
curled up and her wrist twisted.

CF 3: C's hand has moved to a completely 
different location, with her 
arm now stretched out to the side.
\end{Verbatim}

Narration: \emph{``C C fixes the camera"''}
\begin{Verbatim}[commandchars=\\\{\}]
Before: The camera is not 
functioning properly, with 
a loose screw visible.

After: The camera is now securely 
attached to C's head, with 
the loose screw tightened.

CF 1: The camera remains 
loose, still not functioning properly.
CF 2: A makeshift fix is 
attempted, with tape holding
the camera in place."

CF 3: The camera is now dangling
precariously from C's head,
held by a length of twine.
\end{Verbatim}


\subsection{Generated Examples - Video-level}

Summary: \emph{``C was in a room. 
C removed small pieces from a paperwork. 
C constructed a new structure 
with the paperwork pieces.''}

\textcolor{OliveGreen}{Missing-step CF} - 1
\begin{Verbatim}[commandchars=\\\{\}]
C was in a room. 
C constructed a new structure 
with the paperwork pieces.
\end{Verbatim}

\textcolor{OliveGreen}{Missing-step CF} - 2
\begin{Verbatim}[commandchars=\\\{\}]
C was in a room. 
C removed small pieces from 
a paperwork with a tool. #summary
\end{Verbatim}

\textcolor{BrickRed}{Misordered-step CF} - 1
\begin{Verbatim}[commandchars=\\\{\}]
C was in a room.
C constructed a new structure with
the paperwork pieces. 
C removed small pieces from the 
paperwork to create the craft.
\end{Verbatim}

\textcolor{BrickRed}{Misordered-step CF} - 2
\begin{Verbatim}[commandchars=\\\{\}]
C was in a room. 
C removed small pieces from 
the paperwork with a tool. 
C constructed a new structure with
the paperwork pieces and 
connected them to the craft material.
\end{Verbatim}

Summary: \emph{``C attached wooden mechanical 
model pieces together at a table in a room. 
C detached wooden pieces from puzzle sheets. 
C studied an instruction manual''}

\textcolor{OliveGreen}{Misordered-step CF} - 1
\begin{Verbatim}[commandchars=\\\{\}]
C attached wooden mechanical model
pieces together at a table in a room. 
C picks up the puzzle sheet on 
the cutting sheet. 
C studied an instruction manual
\end{Verbatim}

\textcolor{OliveGreen}{Misordered-step CF} - 2
\begin{Verbatim}[commandchars=\\\{\}]
C picks the laser cut wood craft kit. 
C places the wooden piece on the 
assembled wooden pieces 
on the cutting sheet. 
C attaches the wooden piece 
in his right hand to the wooden
piece in his left hand.
\end{Verbatim}

\textcolor{BrickRed}{Misordered-step CF} - 1
\begin{Verbatim}[commandchars=\\\{\}]
C was in a room. 
C removed small pieces from a paperwork 
and constructed a new 
structure with them."
\end{Verbatim}

\textcolor{BrickRed}{Misordered-step CF} - 2
\begin{Verbatim}[commandchars=\\\{\}]
C was in a room.
C constructed a new structure with 
the paperwork pieces.
\end{Verbatim}






\subsection{Quality of Generated Descriptions}
We asked 22 human raters and Gemini 2.5 Pro to evaluate 300 and 1000 pairs of LLM-generated state changes (SC) and their counterfactuals (CF), respectively, by Likert-scoring from 1 to 5 for \emph{Relevance} ($R$) and \emph{Plausibility} ($P$). 
\textbf{Human scores}: SC$_{R}$: 4.95, CF$_{R}$: 4.84; SC$_{P}$: 4.73, CF$_{P}$: 3.87. \textbf{Gemini scores}: SC$_{R}$: 4.85, CF$_{R}$: 4.58 ; SC$_{P}$: 4.91, CF$_{P}$: 4.57. Despite being reasonable and relevant, we found that the generated CFs occasionally reflect low-probability scenarios, suggesting a tradeoff between creativity
and realism in LLMs.
Yet, the ablations in 
the main paper
verify their effectiveness and robustness on procedure-aware tasks.

%% file: supp_sec/additional_results.tex
\section{Expanded Results}
\label{sec:expanded_results}

\begin{table*}[h!]
\scriptsize
\caption{Action phase classification results on the Align-Ego-Exo dataset~\cite{xue2023learning}. ``All" denotes the average across actions.}
\resizebox{1\linewidth}{!}{
    \begin{tabular}{@{}lc|cccccccc|cc@{}}
        \toprule
        && \multicolumn{2}{c}{{Break Eggs}} & \multicolumn{2}{c}{{Pour Milk}} & \multicolumn{2}{c}{{Pour Liquid}} & \multicolumn{2}{c}{{Tennis Forehand}} & \multicolumn{2}{c}{{All}} \\
        {Method} & Pretraining Data & ego+exo & ego & ego+exo & ego & ego+exo & ego & ego+exo & ego & ego+exo & ego \\
        \midrule
        CLIP~\cite{radford2021learning} & WIT~\cite{srinivasan2021wit}+\textbf{\textcolor{red}{Text}} & 50.1 & 54.9 & \underline{50.4} & \textbf{49.8} & 61.3 & 63.7 & \textbf{76.3} & \textbf{78.2} & \underline{59.5} & 61.6\\
        MIL-NCE~\cite{miech2020end} & HTM & 45.5 & 45.0 & 45.9 & 44.2 & 61.2 & 65.3 & 59.5 & 62.3 & 53.0 & 54.2\\
        PVRL~\cite{zhong2023learning} & HTM & \underline{54.6} & \underline{60.6} & \textbf{51.6} & 46.6 & \underline{63.0} & \underline{69.0} & 68.2 & 74.5 & 59.4 & \underline{62.7}\\
        \midrule
        Ours & Ego4D & \textbf{56.2} & \textbf{65.8} & 48.1 & \underline{47.6} & \textbf{68.1} & \textbf{70.6} & \underline{72.7} & \underline{75.1} & \textbf{61.3} & \textbf{64.8} \\
        \bottomrule
    \end{tabular}
}
\label{table:ae2_cls}
\end{table*}

\begin{table*}[h!]
\scriptsize
\caption{Action phase frame retrieval results on the Align-Ego-Exo dataset~\cite{xue2023learning}. ``All" denotes the average across actions.}
\resizebox{1\linewidth}{!}{
    \begin{tabular}{@{}lc|cccccccc|cc@{}}
        \toprule
        && \multicolumn{2}{c}{{Break Eggs}} & \multicolumn{2}{c}{{Pour Milk}} & \multicolumn{2}{c}{{Pour Liquid}} & \multicolumn{2}{c}{{Tennis Forehand}} & \multicolumn{2}{c}{{All}} \\
        {Method} & Pretraining Data & ego+exo & ego & ego+exo & ego & ego+exo & ego & ego+exo & ego & ego+exo & ego \\
        \midrule
        CLIP~\cite{radford2021learning} & WIT~\cite{srinivasan2021wit}+\textbf{\textcolor{red}{Text}} & \underline{63.5} & \underline{68.0} & \textbf{59.3} & \underline{59.2} & 55.9 & 56.1 & \underline{79.1} & 88.7 & \underline{64.4} & \underline{68.0}\\
        MIL-NCE~\cite{miech2020end} & HTM & 58.0 & 57.4 & 47.3 & 51.0 & \underline{57.7} & \underline{59.2} & 74.8 & 84.3 & 53.0 & 54.2\\
        PVRL~\cite{zhong2023learning} & HTM & 59.5 & 63.1 & \underline{58.3} & \textbf{59.3} & 50.2 & 55.1 & 78.3 & \textbf{88.9} & 61.6 & 66.3\\
        \midrule
        Ours & Ego4D & \textbf{66.5} & \textbf{69.4} & 51.4 & 54.9 & \textbf{62.4} & \textbf{67.8} & \textbf{79.4} & \textbf{88.9} & \textbf{64.9} & \textbf{70.3} \\
        \bottomrule
    \end{tabular}
}
\label{table:ae2_retrieval}
\end{table*}

\subsection{Full Results of Action Phase Classification \& Retrieval}
\label{subsec:full}
Tables \ref{table:ae2_cls} and \ref{table:ae2_retrieval} show classification and retrieval results on the Align-Ego-Exo~\cite{xue2023learning} dataset for each action, demonstrating the strong effectiveness of our representations on \textit{short-term} and \textit{fine-grained} procedure awareness.

\subsection{Qualitative Results}
\label{sub}
\begin{figure*}[t!]
\centering
        \includegraphics[width=17cm]{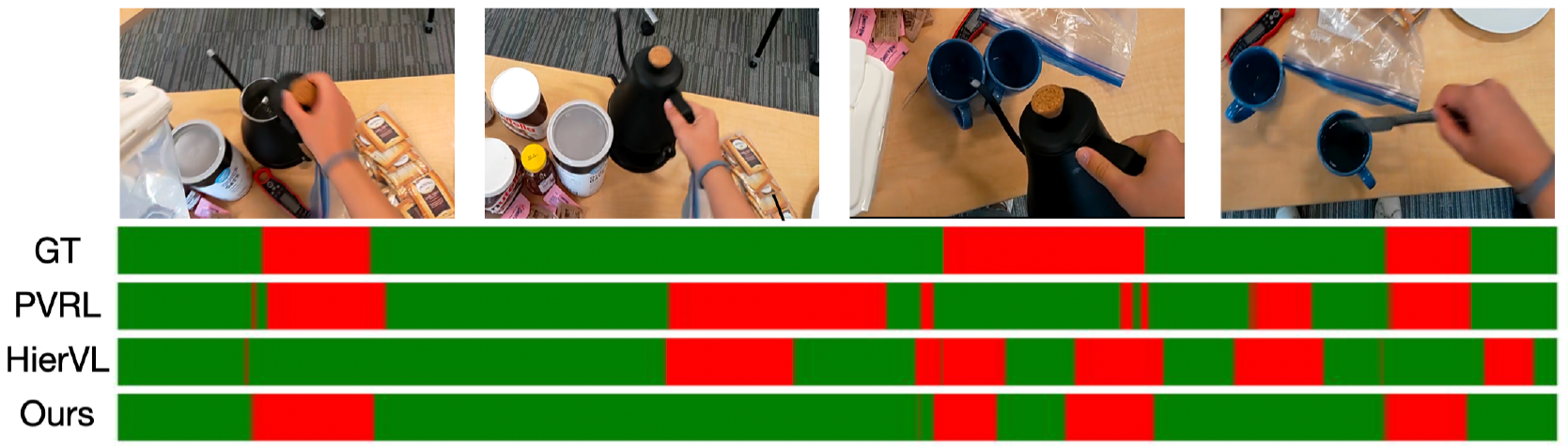}
        \caption{
                 Qualitative results of Error Detection on EgoPER. GT denotes ground truth. \textcolor{OliveGreen}{Green}/\textcolor{red}{Red} segmentation and text denote the normal and error labels, respectively. The presented erroneous procedure \emph{``Make tea''} consists of [\emph{``\textcolor{red}{Not checking water temperature in the kettle},''} \emph{``\textcolor{OliveGreen}{{Hold the kettle,}}'' \emph{\textcolor{red}{``Pour water immediately,''} }\emph{\textcolor{red}{``Stir with the knife'',}} ....}].
                 } 
        
        \label{fig:qualitative}
        \vspace{4mm}
\end{figure*}

\begin{figure*}[t!]
\centering
        \includegraphics[width=17cm]{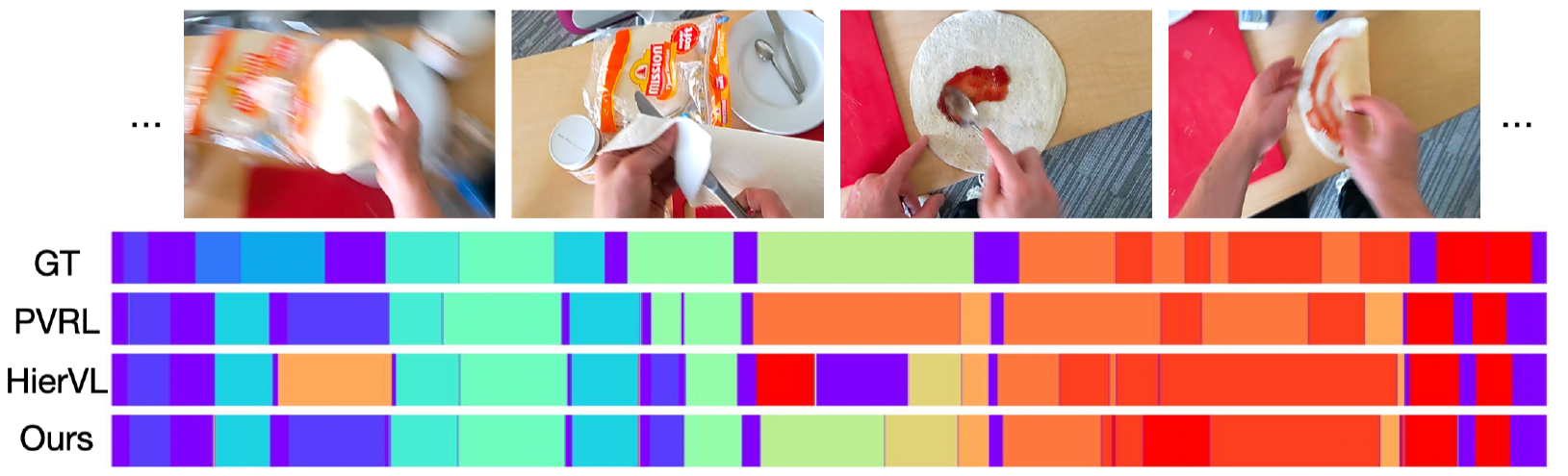}
        \caption{
                Qualitative results of Temporal Action Segmentation on EgoPER. The example presents the procedure \emph{``Make Pinwheel''}. Distinct colored segments are different action step classes.
                 } 
        
        \label{fig:qualitative_2}
        \vspace{3mm}
\end{figure*}

Figure~\ref{fig:qualitative} presents qualitative results on error detection in EgoPER, showing that our learned representations identify erroneous activities with greater fidelity. PVRL and HierVL produce several false positives, leading to over-segmentation, whereas our method better aligns with the ground-truth segments in both temporal location and count. 
Figure~\ref{fig:qualitative_2} shows qualitative results on Temporal Action Segmentation, where PVRL and HierVL misclassify large temporal segments (dark and light orange, respectively). In contrast, our model more accurately distinguishes and classifies actions, reflecting an improved understanding of procedural activities and aligning with our quantitative results.